# Does a Plane Imitate a Bird?
# Does Computer Vision Have to Follow Biological Paradigms?

Emanuel Diamant

VIDIA-mant, P.O. Box 933, 55100 Kiriat Ono, Israel
`emanl@012.net.il`

**Abstract.** We posit a new paradigm for image information processing. For the last 25 years, this task was usually approached in the frame of Triesman's two-stage paradigm [1]. The latter supposes an unsupervised, bottom-up directed process of preliminary information pieces gathering at the lower processing stages and a supervised, top-down directed process of information pieces binding and grouping at the higher stages. It is acknowledged that these sub-processes interact and intervene between them in a tricky and a complicated manner. Notwithstanding the prevalence of this paradigm in biological and computer vision, we nevertheless propose to replace it with a new one, which we would like to designate as a two-part paradigm. In it, information contained in an image is initially extracted in an independent top-down manner by one part of the system, and then it is examined and interpreted by another, separate system part. We argue that the new paradigm seems to be more plausible than its forerunner. We provide evidence from human attention vision studies and insights of Kolmogorov's complexity theory to support these arguments. We also provide some reasons in favor of separate image interpretation issues.

## 1  Introduction

It is generally acknowledged that our computer vision systems have been and continue to be an everlasting attempt to imitate their biological counterparts. As such, they have always faithfully followed the ideas and trends borrowed from the field of biological vision studies. However, image information processing and image understanding issues have remained a mystery and a lasting challenge for both of them. Following biological vision canons, prevalent computer vision applications apprehend image information processing as an interaction of two inversely directed sub-processes. One is – an unsupervised, bottom-up evolving process of low-level elementary image information pieces discovery and localization. The other – is a supervised, top-down propagating process, which conveys the rules and the knowledge that guide the linking and grouping of the preliminary disclosed features into more large agglomerations and sets. It is generally believed that at some higher level of the processing hierarchy this interplay culminates with the required scene decomposition (segmentation) into its meaningful constituents (objects).

As said, the roots of such an approach are easily traced to the Treisman's Feature Integrating Theory [1], Biederman's Recognition-by-components theory [2], and Marr's theory of early visual information processing [3]. They all shared a common belief that human's mental image of the surrounding is clear and full, and point by point defined and specified. On this basis, a range of bottom-up proceeding techniques has been developed and continues to flourish. For example, super-fast Digital Signal Processors (DSPs) with Gigaflop processing power, which were designed to cope with input data inundation. Or Neural Nets that came to solve the problems of data patterns discovery, learned and identified in massive parallel processing arrangements. Or the latest wave of computational models for selective attention vision studies [4].

With only a minor opposition [5], the bottom-up/top-down processing principle has been established as an incontestable and dominating leader in both biological and computer vision.

## 2   Denying the Two Stage Approach

The flow of evidence that comes from the latest selective attention vision studies encourages us to reconsider the established dogmas of image processing. First of all, the hypothesis that our mental image is entirely clear and crisp does not hold more, it was just an inspiring illusion [6]. In the last years, various types of perceptual blindness have been unveiled, investigated and described [7].

Considering selective attention vision studies, it will be interesting to note that the latest investigations in this field also come in contradiction with the established bottom-up/top-down approaches. After all, it was a long-standing conviction that the main part of the incoming visual information is acquired via the extremely dense populated (by photoreceptors) eye's part called fovea. Because of its very small dimensions, to cover the entire field of view, the eyes constantly move the fovea, redirecting the gaze and placing the fovea over different scene locations, thus enabling successful gathering of the required high-resolution information. A more scrutinizing view on the matters reveals that the decision to make the next saccadic move precedes the detailed information gathering performed at such a location. That leads to an assumption that other sorts of information must be involved, supporting attention focusing mechanisms.

Considering the empirical evidence (and the references that we provide are only a negligible part of an ample list of recent publications), juxtaposing it with the insights of Kolmogorov Complexity theory (which we adopt to explain these empirical biological findings), we have come to a following conclusion: the bottom-up/top-down principle can not be maintained any more. It must be replaced with a more suitable approach.

Recently, we have published a couple of papers ([8], [9]) in which we explain our view on the issue. For the clarity of this discussion, we will briefly repeat their main points. First, we reconsider the very notion of image information content. Despite of its widespread use, the notion of it is still ill defined, intuitive, and ambiguous. Most often, it is used in the Shannon's sense, which means information content assessment averaged over the whole signal ensemble (an echo of the bottom-up approach).

Humans, however, rarely resort to such estimates. They are very efficient in decomposing images into their meaningful constituents and then focusing attention to the most perceptually important and relevant image parts. That fits the concepts of Kolmogorov's complexity theory, which explores the notions of randomness and information. Following the insights of this theory, we have proposed the next definition of image contained information: image information content can be defined as a set of descriptions of the visible image data structures. Three levels of such description can be generally distinguished: 1) the global level, where the coarse structure of the entire scene is initially outlined; 2) the intermediate level, where structures of separate, non-overlapping image regions usually associated with individual scene objects are delineated; and 3) the low level description, where local image structures observed in a limited and restricted field of view are resolved.

The Kolmogorov Complexity theory prescribes that the descriptions must be created in a hierarchical and recursive manner, that is, starting with a generalized and simplified description of image structure, it proceeds in a top-down fashion to more and more fine information details elaboration performed at the lower description levels.

A practical algorithm, which implements this idea, is presented, and its schema is depicted in the Figure 1.

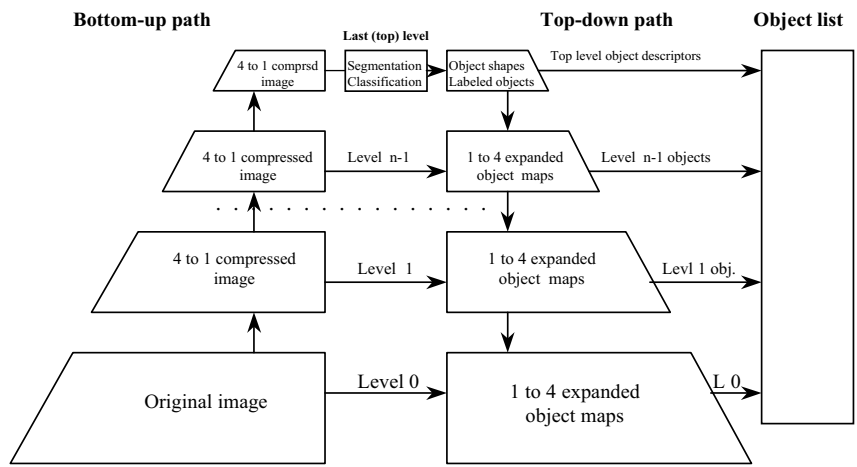

**Fig. 1.** The Schema of the proposed approach

As it can be seen from the figure, the schema is comprised of three processing paths: the bottom-up processing path, the top-down processing path and a stack where the discovered information content (the generated descriptions of it) are actually accumulated.

As it follows from the schema, the input image is initially squeezed to a small size of approximately 100 pixels. The rules of this shrinking operation are very simple and fast: four non-overlapping neighbour pixels in an image at level $L$ are averaged and the result is assigned to a pixel in a higher $(L+1)$-level image. This is known as "four children to one parent relationship". Then, at the top of the shrinking pyramid, the image is segmented, and each segmented region is labeled. Since the image size at the top is significantly reduced and since in the course of the bottom-up image squeezing a severe data averaging is attained, the image segmentation/classification procedure does not demand special computational resources. Any well-known segmentation methodology will suffice. We use our own proprietary technique that is based on a low-level (local) information content evaluation, but this is not obligatory.

From this point on, the top-down processing path is commenced. At each level, the two previously defined maps (average region intensity map and the associated label map) are expanded to the size of an image at the nearest lower level. Since the regions at different hierarchical levels do not exhibit significant changes in their characteristic intensity, the majority of newly assigned pixels are determined in a sufficiently correct manner. Only pixels at region borders and seeds of newly emerging regions may significantly deviate from the assigned values. Taking the corresponding current-level image as a reference (the left-side unsegmented image), these pixels can be easily detected and subjected to a refinement cycle. In such a manner, the process is subsequently repeated at all descending levels until the segmentation/classification of the original input image is successfully accomplished.

At every processing level, every image object-region (just recovered or an inherited one) is registered in the objects' appearance list, which is the third constituting part of the proposed scheme. The registered object parameters are the available simplified object's attributes, such as size, center-of-mass position, average object intensity and hierarchical and topological relationship within and between the objects ("sub-part of…", "at the left of…", etc.). They are sparse, general, and yet specific enough to capture the object's characteristic features in a variety of descriptive forms.

Finally, it must be explicitly restated: all this image information content discovery, extraction and representation proceeds without any involvement of any high-level knowledge about semantic nature of an image or any cognitive guidance cues mediating the process. However, that does not preclude a human observer to grasp the gist of the segmented scene in a clear and unambiguous way. (Which confirms that all information needed for gist comprehension is extracted and is represented correctly.)

## 3   Illustrative Example

To illustrate the qualities of the image information extraction part we have chosen a scene from the Photo-Gallery of the Natural Resources Conservation Service, USA Department of Agriculture, [10].

Figure 2 represents the original image, Figures 3 – 7 illustrate segmentation results at various levels of the processing hierarchy. Level 5 (Fig. 3) is the topmost nearest level (For the image of this size the algorithm has created a 6-level hierarchy). Level 1 (Fig. 7) is the lower-end closest level. For space saving, we do not provide all the samples of the segmentation succession, but for readers' convenience each presented example is expanded to the size of the original image.

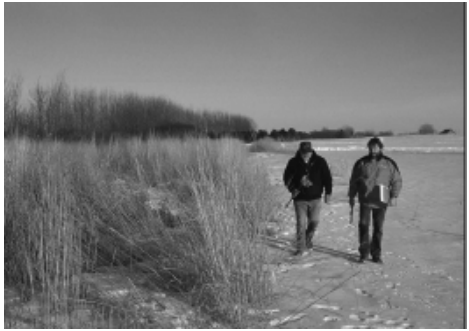

**Fig. 2.** Original image, size 1052x750 pixels

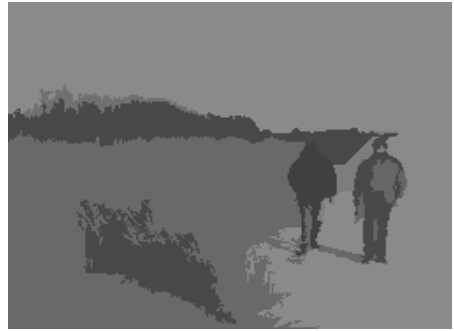

**Fig. 3.** Level 5 decompos., 8 region-objects

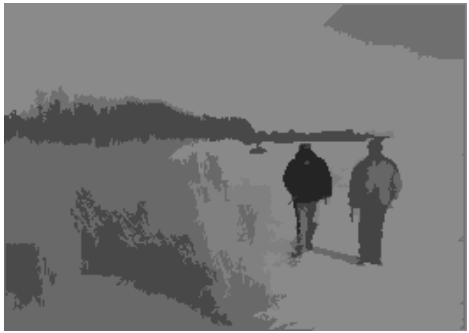

**Fig. 4.** Level 4 decompos., 14 region-objects

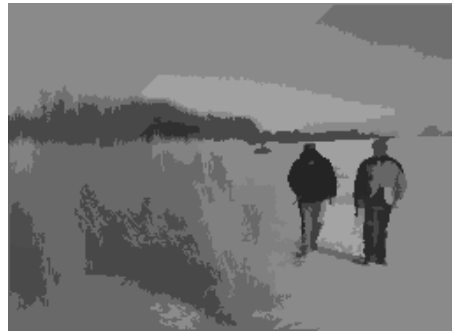

**Fig. 5.** Level 3 decompos., 27 region-objects

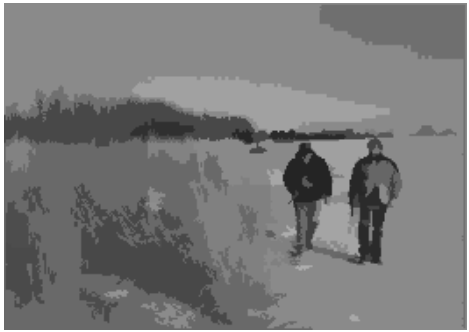

**Fig. 6.** Level 2 decompos., 49 region-objects

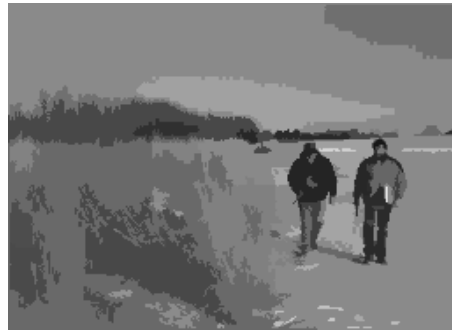

**Fig. 7.** Level 1 decompos., 79 region-objects

Extracted from the object list, numbers of distinguished (segmented) at each corresponding level regions (objects) are given in each figure capture.

Because real object decomposition is not known in advance, only the generalized intensity maps are presented here.

## 4   Introducing Image Interpretation

Eliminating information content extraction from the frame of the bottom-up/top-down approach and declaring its independent, self-consistent and unsupervised top-down manner of information processing, immediately raises a question: and what is about high-level cognitive image perception? Indeed, none at any time has ever denied the importance of cognitive treatment of image content. But the autonomous nature of image information content preprocessing (that we have just above defined and approved) does not leave any choices for an anticipated answer: understanding of image information content, that means, its appropriate interpretation, must come from the outside, from another part of the processing system. Contrary to the bottom-up/top-down approach, this part has no influence on its predecessor.

The consequences of acceptance of such a two-part processing concept are tremendous. First of all, the common belief that the knowledge needed for high-level information processing can be learned from the input data itself is totally invalidated. Now, all of the so cherished training and learning theories, neural nets and adaptive approximators – all that must be put in junk. And then... Regarding image interpretation duties (the functionality of the second system's part), several questions must be urgently considered: 1) how the knowledge, packed into a knowledge base that supports the interpretation process, is initially acquired? How and from where does it come? 2) how it must be presented? What is the best representation form of it? 3) how the interaction with the information content (the image stuff subjected to interpretation and contained in the preceding system's module) is actually performed?

We hope that we have the right answers. At least, we will try to put them unambiguously. For the first question, we think that the knowledge must come from the system designer, from his image context understanding and his previous domain-related experience. As in humans, the prime learning and knowledge accumulation process must be explicit and declarative. That means, not independently acquired, but deliberately introduced. As in humans, the best form for such introduction, its further memorization for later recall, its representation and usage – is an ontology [11]. (And that is the answer for the second question.) By saying this, we don't mean the world's ontology that a human gradually creates in his life span. We mean a simplified, domain-restricted and contextualized ontology, or as it is now called – domain interpretation schema [12]. Which can be very specific about image information content and context, and does not have to share knowledge with other applications. This makes it very flexible, easily designed by the application supervisor, which thus becomes a single source for both the required knowledge and its representation in a suitable form (of an interpretation schema).

A known way to avoid complications in ontology maintenance and updating (in accordance with the changing application environment) is to create additional partial interpretation schemas, which take into account the encountered changes. To make the whole system workable, a cross mapping between partial schemas must be established. Such mapping is a part of a local representation, and, as we see that, must be also provided by the system designer. However, he has not to do this in advance, he can gradually expand and increase the system's interpretation abilities adding new ontologies as the previous arrangement becomes insufficient.

Finally, and that is the first time when the idea is announced, we propose to see the description list at the output of the first module (the early described information processing module) as a special kind of a partial ontology, written in a special description language. By the way, this language can be shared with attribute description languages utilized in the partial ontologies. Once more, providing the mapping between them paves the way for the whole system integration. And that is the answer for the third question.

The proposed framework does not solve the whole image interpretation problem. It must be seen only as a first step of it, where segmented in an unsupervised manner image regions become meaningfully regrouped and bonded into human accustomed objects with human familiar lexical names and labels. The latter can be used then in further more advanced interpretations of image spatio-temporal content.

## 5   Conclusions

In this paper, we have presented a new paradigm for image information content processing. Contrary to the traditional two-stage paradigm, which rely on a bottom-up (resource exhaustive) processing and on a top-down mediating (which requires external knowledge incorporation), our paradigm assumes a two-part approach. Here, one part is responsible for image information extraction (in an unsupervised top-down proceeding manner) and the other part is busy with interpretation of this information. Such subdivision of functional duties more reliably represents biological vision functionality, (albeit, it is still not recognized by biological vision research community).

The two-part paradigm forces reconsideration of many other image information related topics. For example, Shannon's definition of information, as an average over an ensemble, versus Kolmogorov's definition of information, as a shortest program that reliably describes/reproduces the structure of image objects. A new viewpoint must be accepted regarding information interpretation issues, such as knowledge acquisition and learning, knowledge representation (in form of multiple parallel ontologies), and knowledge consolidation via mutual cross-mapping of the ontologies.

A hard research and investigation future work is anticipated. We hope it would be successfully fulfilled.

## References

1. A. Treisman and G. Gelade, "A feature-integration theory of attention", *Cognitive Psychology*, 12, pp. 97-136, Jan. 1980.
2. I. Biederman, "Recognition-by-components: A theory of human image understanding", *Psychological Review*, vol. 94, No. 2, pp. 115-147, 1987.
3. D. Marr, "Vision: A Computational Investigation into the Human Representation and Processing of Visual Information", Freeman, San Francisco, 1982.
4. L. Itti, "Models of Bottom-Up Attention and Saliency", In: *Neurobiology of Attention*, (L. Itti, G. Rees, J. Tsotsos, Eds.), pp. 576-582, San Diego, CA: Elsevier, 2005.
5. D. Navon, "Forest Before Trees: The Precedence of Global Features in Visual Perception", *Cognitive Psychology*, **9**, pp. 353-383, 1977.

6. A. Clark, "Is Seeing All It Seems? Action, Reason and the Grand Illusion", *Journal of Consciousness Studies*, vol. 9, No. 5/6, pp. 181-218, May – June 2002.
7. D. J. Simons and R. A. Rensink, "Change blindness: past, present, and future", *Trends in Cognitive Science*, vol. 9, No. 1, pp. 16 – 20, January 2005.
8. E. Diamant, "Image information content estimation and elicitation", *WSEAS Transactions on Computers*, vol. 2, issue 2, pp. 443-448, April 2003.
9. E. Diamant, "Searching for image information content, its discovery, extraction, and representation", *Journal of Electronic Imaging,* vol. 14, issue 1, article 013016, January-March, 2005.
10. NRCS image collection. Available: http://photogallery.nrcs.usda.gov/ (Iowa collection).
11. M. Uschold and M. Gruninger, "ONTOLOGIES: Principles, Methods and Applications", *Knowledge Engineering Review*, vol. 11, No. 2, pp. 93-155, 1996.
12. P. Bouquet, F. Giunchiglia, F. van Harmelen, L. Serafini, and H. Stuckenschmidt, "C-OWL: Contextualizing Ontologies", *Second International Semantic Web Conference (ISWC-2003)*, LNCS vol. 2870, pp. 164-179, Springer Verlag, 2003.